\documentclass{article} 
\usepackage{iclr2021_conference,times}

\usepackage{amsmath,amsfonts,bm}

\def\eqref#1{equation~\ref{#1}}

\def\1{\bm{1}}

\DeclareMathAlphabet{\mathsfit}{\encodingdefault}{\sfdefault}{m}{sl}
\SetMathAlphabet{\mathsfit}{bold}{\encodingdefault}{\sfdefault}{bx}{n}

\usepackage{hyperref}
\usepackage{url}
\usepackage{bbm,amsmath}

\title{Federated Learning without Revealing the Decision Boundaries}

\author{Roozbeh Yousefzadeh 
\\
Yale University and VA Connecticut Healthcare System\\
\texttt{roozbeh.yousefzadeh@yale.edu} \\
}

\iclrfinalcopy 
\begin{document}

\maketitle

\begin{abstract}
We consider the recent privacy preserving methods that train the models not on original images, but on mixed images that look like noise and hard to trace back to the original images. We explain that those mixed images will be samples on the decision boundaries of the trained model, and although such methods successfully hide the contents of images from the entity in charge of federated learning, they provide crucial information to that entity about the decision boundaries of the trained model. Once the entity has exact samples on the decision boundaries of the model, they may use it for effective adversarial attacks on the model during training and/or afterwards. If we have to hide our images from that entity, how can we trust them with the decision boundaries of our model? As a remedy, we propose a method to encrypt the images, and have a decryption module hidden inside the model. The entity in charge of federated learning will only have access to a set of complex-valued coefficients, but the model will first decrypt the images and then put them through the convolutional layers. This way, the entity will not see the training images and they will not know the location of the decision boundaries of the model.
\end{abstract}

\section{Introduction}

Our focus in this paper is on privacy methods for federated learning and we specifically consider the methods that hide the contents of images from the entity in charge of federated training, denoted by $\mathcal{T}$. In this setting, $\mathcal{T}$ is in charge of training a model, but training images are owned by several entities that are not willing to share their images with each other and with $\mathcal{T}$. Owners are willing to participate in training the model, only if their images are kept private. For example, image owners can be medical institutions that have invested in analyzing and labeling medical images of patients. They may be willing to participate in training an image classification model while keeping their images private \citep{li2020federated}.

Many studies have considered this problem and different methods have been proposed \citep{yang2019federated}. For example, some methods propose that image owners compute the gradient of loss function of the model on their images and provide the gradients to $\mathcal{T}$ instead of providing the images. Several issues may arise in this setting, as the model parameters will change after implementing each gradient batch, and communicating the model parameters to image owners after each update can be quite expensive \citep{konevcny2016federated}.

Another line of research mixes the training images so that they look like noise and then provides those mixed images to $\mathcal{T}$ for training. The method proposed by \citet{huang2020instahide} mixes the images so that they are provably hard to be traced back to original images, and labels the mixed images such that $\mathcal{T}$ can achieve relatively high accuracy when it trains the model on those mixed images. This is the line of research we are considering in this paper. We explain that those mixed images will be exact points on the decision boundaries of the trained model, and although $\mathcal{T}$ does not have access to original training images, it will know where the decision boundaries of the model will be. If we cannot trust $\mathcal{T}$ with our training images, how can we trust it with the decision boundaries of our model?

In the following, we first discuss the decision boundaries of deep learning models. We then explain the limitations of privacy achieved by training methods such as \citet{huang2020instahide}, and finally, we propose the outline of a new method to preserve the privacy of training images without revealing the decision boundaries of the model.

\section{Deep learning models and their decision boundaries}

Deep learning models for image classification are classification functions that partition their domain and assign a class to each partition. Partitions are defined by decision boundaries and so is the model. In concept, this can be viewed the same as a linear regression model. A linear regression model partitions its domain with a hyper-plane where the hyper-plane is its decision boundary. Because the pixel space is bounded and usually normalized, the domain of an image classification model can be considered bounded, too, usually a hyper-cube where each dimension ranges between 0 and~1. The decision boundaries of deep learning models are geometrically complex and far from being a hyper-plane \citep{fawzi2018empirical}, and this complexity is what makes these models powerful classification functions.

As a result, these models/functions are considered black boxes and there are many lingering questions about them \citep{strang2019linear}. The adversarial vulnerability of these models can be explained by the closeness of samples to decision boundaries, but it is not clear why decision boundaries are so close to samples \citep{shafahi2018adversarial}. While many studies try to exploit the decision boundaries of the models for adversarial purposes \citep{tramer2020adaptive}, another set of studies are trying to push the decision boundaries away from samples in order to make the models less vulnerable \citep{cohen2019certified}. Studies on generalization of these models are also concerned about the decision boundaries \citep{elsayed2018large,marginbased2019}.

The training process of image classification models can be viewed as defining the location of decision boundaries in the domain. Recently, there have been some studies that train the models in interesting ways other than just minimizing a loss function. One of the most interesting of these methods is by \citet{zhang2018mixup}. They use conventional training regimes and cross-entropy loss, but instead of training the model on original training images, they mix the images (with different labels) by averaging them pairwise in the pixel space and averaging their labels, as well. This way, they train the models on the mixed images and it turns out to be very effective for better generalization and more robustness of models against adversarial attacks and memorization. Although their study is not focused on the decision boundaries, their training method directly defines the location of decision boundaries in between training samples because the labels they use for mixed images are actually flip labels between classes. 

Later, several studies leveraged the mixing aspect of this method for federated learning \citep{fu2019mixup,boulemtafes2020review}. Building on that line of research, \citet{huang2020instahide} proposed a method called \textit{InstaHide}, where the entity in charge of training the model has access to certain mixed images that cannot be easily traced back to original images in the private training sets. Again, when the model learns such images, it learns where the location of its decision boundaries should be.

\section{The extent of privacy we can achieve}

The method proposed by \citet{huang2020instahide} successfully hides the original images from $\mathcal{T}$. In fact, images used for training do not look anything similar to the original images and it is indeed hard to trace them back to original images. But to what extent does their method help us with preserving privacy?

The images used for training have flip labels which exactly define where the location of decision boundaries of the model should be. It follows that the entity that trains the model has direct access to samples on the decision boundaries of the model, because each training sample is a point on a decision boundary. Hence, by putting the training samples together, that entity can acquire a good picture of the partitions of each class defined in the domain.

If we have to hide the contents of original training samples from the entity in charge of training, should not we be concerned about the entity having access to the decision boundaries of the model? What is the extent of the privacy preserved by hiding the contents of images but giving out samples on the decision boundaries of our model?

From the adversarial attacks literature, we know that when an entity knows the decision boundaries of our models, they will know the functional behavior of it and they can affect that adversarially, during training and also after the training.

\section{Decryption module inside the model}

As a remedy, here we propose a method to encrypt the images with shearlet systems at the origin, so that only shearlet coefficients are shared with entity $\mathcal{T}$. There will be a decryption module inside the model that would decrypt the images before putting them through the convolutional layers. After the model is trained, one can remove the decryption module from the model and directly feed images to its convolutional layers.\footnote{If desired, decryption module can still remain inside the model for the testing time as well.} This way, the model is trained on images, but training images are kept hidden from $\mathcal{T}$. Moreover, $\mathcal{T}$ will have no immediate knowledge about the decision boundaries of the trained model.

In the following, we will briefly explain the shearlet systems which are recently used for image encryption. Using a complicated function to build a shearlet system and using truncated filter boundaries will lead to complex-valued shearlet coefficients, and it would be hard to trace those coefficients back to the actual images without having the shearlet system.

\subsection{Background on Shealets}

Shearlets are a multi-scale framework for efficient processing and representation of multidimensional data, especially images. Shearlets were built on the scientific knowledge of wavelets in order to address the limitations of wavelets in handling anisotropic features, e.g., edge features in images \citep{labate2005sparse}.

Over the decades, special tools have been developed to process images (along with other forms of data and functional representations) and to extract useful features from them. Such tools include the Fourier transform, wavelets, shearlets, etc. In fact, wavelets were developed building on the knowledge of Fourier transform in the context of image and signal processing. For example, \citet{daubechies1990wavelet} showed that wavelets perform better than windowed Fourier transform on visual signals, because wavelets handle the frequencies in a nonlinear way. Although wavelets have proven to be more capable than the Fourier transform, they still have shortcomings when it comes to analyzing edge features, because they can only handle point-wise singularities. Shearlets are well-equipped in this respect because, as their name suggests, they extract anisotropic features by deploying a shear matrix. Interaction between the shear matrix and appropriate scaling allows them to operate on a plane in a rotation-like manner, capturing edge features with various shapes \citep{kutyniok2012shearlets}.

This makes the shearlets more capable in image processing and at the same time, it makes them computationally more sophisticated and therefore, more suitable for encryption. Wavelets have been used for image encryption in a variety of settings \citep{chen2005optical,chen2008image,luo2015symmetrical}, which hints that shearlets can be even more effective in hiding the contents of images, as already used by \citet{kumar2018qr,chen2020generalized}.

\subsection{Shearlet Systems}

In order to decompose images with shearlets, we first need to create a shearlet system. A shearlet system is composed of three operations: Dilation, Shearing, and Translation. Here, we briefly review the definition provided by \citet{andrade2020shearlets}.

Let $A_a, a \in \mathbbm{R} := \mathbbm{R} \backslash \{0\}$ be a periodic scaling matrix and $S_s, s \in \mathbbm{R}$ be a shearing matrix given by

\[
    A_a = \begin{pmatrix}
    a & 0\\
    0 & |a|^{1/2}
    \end{pmatrix} \; \text{and} \;
    S_s = \begin{pmatrix}
    1 & s\\
    0 & 1
    \end{pmatrix}.
\]

For $B \in \mathbbm{R}^{2\times 2}$, the dilation operator is defined by

$$D_B: L^2(\mathbbm{R}^2) \rightarrow L^2(\mathbbm{R}^2) \quad , \quad (D_B f)(x) \mapsto |\det(B)|^{-1/2} f(M^{-1}x)$$

Plugging $A_a$ and $S_s$ as $B$ in the dilation operator yields the scaling and shearing operators, respectively. 
The last ingredient is the translation operator, defined for $T_t, t \in \mathbbm{R}^2$ as

$$T_t: L^2(\mathbbm{R}^2) \rightarrow L^2(\mathbbm{R}^2) \quad , \quad (T_t f)(x) \mapsto f(x-t)$$

Then, for a given function $\psi \in L^2(\mathbbm{R}^2)$ the continuous shearlet system $\mathcal{SH}(\psi)$ is defined by
\begin{equation}
    \mathcal{SH}(\psi) = \{ \psi_{a,s,t} := T_t D_{S_s} D_{A_a} \psi = a^{-3/4} \psi(A_a^{-1} S_s^{-1}(.-t)) : a \in \mathbbm{R}^*, s \in \mathbbm{R}, t \in \mathbbm{R}^2 \}.
\end{equation}


There are certain admissibility requirements for $\psi$. The $\psi$ would actually be the key that would passed to the hidden module in the model and also shared with image owners. Both the hidden module and the image owner will first construct the shearlet system based on the given $\psi$, and then use the $\mathcal{SH}(\psi)$ to decompose or reconstruct the images, as we explain in the next section.


\subsection{Processing images with shearlets}

To decompose image $x$ with shearlet system $\mathcal{SH}(\psi)$, we use
\begin{equation}
    [c] = \mathrm{shdec}(x,\mathcal{SH}(\psi)),
\end{equation}

This operation is reversible, meaning that given a shearlet system and a set of shearlet coefficients, we can reconstruct an image
\begin{equation}
    [x] = \mathrm{shrec}(c,\mathcal{SH}(\psi)).
\end{equation}

The first function will be used by image owners to decompose their images, and the second function will be hidden inside the model to reconstruct the images before putting them through the convolutional layers.

Generally, shearlet coefficients can be real-valued or complex. We suggest using truncated boundary conditions which would lead to complex shearlet coefficients, making it harder for the adversary to decrypt the images.


\section{Conclusion}

In this paper, we reviewed the limitations of privacy preserving methods that mix training images in order to hide them from the entity in charge of federated training. We explained that although such methods hide the contents of training images, they provide exact samples on the decision boundaries of the trained to that same entity. If the entity cannot be trusted with our training images, how can we trust it with the decision boundaries of our models? As a remedy, we proposed using shearlets to encrypt the images while transmitting them in the federated learning system, and then having a decryption module hidden inside the model in order to reconstruct the images before feeding them through the convolutional layers. In this setting, not only we have hidden the training images, we have also hidden the location of the decision boundaries.

\section{Future work}

We shall focus on the choice of $\psi$ for creating the shearlet system and rigorously show that inferring the images from their complex-valued shearlet coefficients will be hard. There has been some studies on this front in the literature which we aim to expand on. It can be particularly useful to study the shearlet encryption in the Shamir's secret sharing scheme \citep{yang2004t}. This has been studied to some extent for wavelets \citep{ulutas2011medical}, but shearlet systems can be more complicated than wavelets.

\subsubsection*{Acknowledgments}
R.Y. is supported by a fellowship from the Department of Veterans Affairs. The views expressed in this manuscript are those of the author and do not necessarily reflect the position or policy of the Department of Veterans Affairs or the United States government.

\bibliography{main}
\bibliographystyle{iclr2021_conference}



\end{document}